\documentclass[conference]{IEEEtran}
\IEEEoverridecommandlockouts
\usepackage{cite}
\usepackage{amsmath,amssymb,amsfonts}
\usepackage{algorithmic}
\usepackage{graphicx}
\usepackage{textcomp}
\usepackage{xcolor}
\def\BibTeX{{\rm B\kern-.05em{\sc i\kern-.025em b}\kern-.08em
    T\kern-.1667em\lower.7ex\hbox{E}\kern-.125emX}}
\begin{document}

\title{How to Train an Accurate and Efficient Object Detection Model on Any Dataset}

\author{\IEEEauthorblockN{Galina Zalesskaya}
\IEEEauthorblockA{\textit{Intel} \\
Israel \\
galina.zalesskaya@intel.com }

\and
\IEEEauthorblockN{Bogna Bylicka}
\IEEEauthorblockA{\textit{Intel} \\
Poland \\
bogna.bylicka@intel.com }

\and
\IEEEauthorblockN{Eugene Liu}
\IEEEauthorblockA{\textit{Intel} \\
United Kingdom \\
eugene.liu@intel.com }
}

\maketitle

\begin{abstract}

The rapidly evolving industry demands high accuracy of the models without the need for time-consuming and computationally expensive experiments required for fine-tuning. Moreover, a model and training pipeline, which was once carefully optimized for a specific dataset, rarely generalizes well to training on a different dataset. This makes it unrealistic to have carefully fine-tuned models for each use case. To solve this, we propose an alternative approach that also forms a backbone of Intel{\textregistered} Geti{\texttrademark} platform: a dataset-agnostic template for object detection trainings, consisting of carefully chosen and pre-trained models together with a robust training pipeline for further training. Our solution works out-of-the-box and provides a strong baseline on a wide range of datasets. It can be used on its own or as a starting point for further fine-tuning for specific use cases when needed. We obtained dataset-agnostic templates by performing parallel training on a corpus of datasets and optimizing the choice of architectures and training tricks with respect to the average results on the whole corpora. We examined a number of architectures, taking into account the performance-accuracy trade-off. Consequently, we propose 3 finalists, \textbf{VFNet}, \textbf{ATSS}, and \textbf{SSD}, that can be deployed on CPU using the OpenVINO{\texttrademark} toolkit. The source code is available as a part of the OpenVINO{\texttrademark} Training Extensions\footnote{https://github.com/openvinotoolkit/training\_extensions}.

\end{abstract}

\IEEEpeerreviewmaketitle

\section{Introduction}

Over recent years, deep learning methods have been increasingly used for Computer Vision tasks, including Object Detection.
While the most of the architectures are optimised on well known benchmarks like COCO \cite{coco} \cite{yolov4} \cite{dino}, further remarkable results have been obtained using CNNs for domain-specific tasks, such as medical \cite{medical_OD}, traffic \cite{traffic}, environmental \cite{fire} etc. Such domain specific solutions, however, are usually heavily optimized for a specific target dataset, starting from carefully tailored architecture to choice of training tricks. This not only makes them time-consuming and computationally heavy to obtain but also is the reason why they \textbf{lack good domain generalization}. Training models in such a way suffers from the caveat of overly adjusting the techniques to a specific dataset. That leads to, on one hand, great predictions for a given dataset, on the other, showing poor transferability to a different one. Meaning, such a model is unlikely to train well on a new, previously unseen dataset, without additional time-consuming and computationally expensive experiments to find the optimal tweaks of the model architecture and training pipeline. Hence, such an approach, even though used in the past with great successes, is \textbf{not flexible enough} for the rapidly evolving needs across a wide variety of computer vision use cases with a focus on time to value.  

In this paper we propose an alternative approach aimed at those situations, addressing the needs of fast-paced industry and real-world applications. We provide a \textbf{dataset-agnostic template for object detection that is ready to be used straight away} and guarantees a strong baseline on a wide variety of datasets, without needing additional time-consuming and computationally demanding experiments. The template also provides the flexibility for further fine-tuning, however, it is not required. This methodology includes a carefully chosen and pre-trained model, together with a robust and efficient training pipeline for further training. 

To achieve this, we base this study on a carefully chosen corpus of 11 datasets that cover a wide range of domains and sizes. They differ in both number of images as well as their complexity, including the number of classes, object sizes, distinguishability from the background, and resolution.
The training procedure involves parallel experiments on all the datasets from the corpus, optimizing the model with respect to average metrics over all corpora.


To summarize, we introduce:
\begin{itemize}
\item a training paradigm together with a choice of a corpus of datasets,
\item a universal, dataset-agnostic model template for object detection training,
\item additional patience parameters for ReduceOnPlateau and Early Stopping, to correct for the different sizes of datasets and avoid over/under-fitting,
\item choice of data-agnostic augmentations and specific tricks for each of our 3 models that work on a wide range of datasets
\end{itemize}

This methodology in the Intel{\textregistered} Geti{\texttrademark} computer vision platform enables platform users to accelerate their object detection training workloads for their real-world use cases across a variety of industries.
\section{Related Work}

\subsection{Architectures}
In this section, we will briefly describe the architectures we explore in our experiments.

\textbf{SSD}
Since its introduction, SSD \cite{SSD} has been a go-to model in the industry, thanks to its high computational efficiency. As an anchor-based detector, it predicts detections on the predefined anchors. All generated predictions are classified as positive or negative, and then an extra offset regression on the positives is done for the final refinement of the bounding box location. However, anchor-based models present a serious disadvantage, they need manual tuning of the number of anchors, their ratio, and scale. This fact negatively affects their ability to generalize to different datasets.

\textbf{FCOS} Anchor-free detectors address these concerns. Instead of anchor boxes, they use single points for detection, determining the center of the object. FCOS \cite{FCOS} approaches object detection in a per-pixel way, making use of the multi-level predictions thanks to FPN. For each layer of the feature pyramid, each point on the feature map can generate a prediction, that is later classified as positive or negative, and further regression is done to determine the bounding box around the initial point. Since the predictions are made on points, a new way to define positives and negatives is introduced to replace the traditional IoU-based strategy. This brings further improvements.  Additionally, to suppress the low-quality detections during NMS, \textit{centerness}, evaluating localization estimation, is used in combination with classification score as a new ranking measure.

\textbf{YOLOX}  YOLOX \cite{YOLOX} is a new anchor-free member of another classic family of models. As a basis it takes YOLOv3 \cite{YOLOv3}, still widely used in the industry, then boosting it for modern times with the newest tricks, including decoupled head, to deal with classification VS regression conflict, and advanced label assigning strategy, SimOTA.

\textbf{ATSS}  ATSS \cite{ATSS} improves performance of detection, due to introduction of \textit{Adaptive Training Sample Selection}. It is a method that allows for an automatic selection of positive and negative samples based on the statistical characteristics of objects.

\textbf{VFNet}
VFNet \cite{VFNet} builds on FCOS and ATSS advances, topping it up with a new ranking measure for the NMS stage, to suppress low-quality detections even more efficiently. The proposed metric, \textit{the IoU-aware classification score}, combines together information on the classification score and localization estimation in a more natural way and in only one variable. This not only gives better results in terms of the quality of detections but also avoids the overhead of the additional branch for localization score regression.

\textbf{Faster R-CNN and Cascade R-CNN}
All the models considered so far were one-stage detectors.
Faster R-CNN \cite{faster} is a two-stage detector, where in the first stage the model coarse scans the whole image to generate region proposals, on which it concentrates in more detail in the second stage. A Region Proposal Network is introduced for the effective generation of region proposals. It is an anchor-based detector with a definition of positive and negative samples through IoU threshold. This generates issues with the quality of detections, low threshold produces noisy detections, increased threshold causes vanishing positive samples and effectively overfitting. Cascade R-CNN \cite{cascade} is a multi-stage detector addressing this issue. It consists of a sequence of detectors trained with increasing IoU thresholds.

\subsection{Towards More Universal Models}
Recently a lot of work has been done in the area of moving towards more flexible, generalizable models in computer vision.
This ranges from multi-task training \cite{multi_task}, where a model is simultaneously trained to perform different tasks,  to domain generalization \cite{dom_gen}, where the goal is to build models that can predict well on datasets out-of-distribution from those seen during training. Another interesting direction is a few-shot dataset generalisation  \cite{few_shot_gen}.
In this approach parallel training on diverse datasets is used to build a universal template that can adapt to unseen data using only a few examples.
Similar work to ours has been done in the classification domain \cite{custom_class}, where adaptive training strategies are proposed and proved to be working by training lightweight models on a number of diverse image classification datasets.

\section{Method}
This section describes in detail our approach to train dataset-agnostic models from the architecture to the training tricks.

\subsection{Architecture}
To create a scope of the models that are applicable for the various object detection datasets regardless of difficulty and object size, we experimented with architectures in three categories: the light-weighted (about 10 GFLOPs), highly accurate, and medium ones. A detailed comparison of the different architectures used can be found in Table~\ref{tab:avg_metrics}.

For the fast and medium models we took light and accurate enough MobileNetV2 backbone \cite{mobilenetv2}, which is commonly used in the application field.
For the accurate model we used ResNet50 \cite{resnet} as a backbone.

For the light model we used really fast SSD with 2 heads, which can be even more optimized for CPU using OpenVINO{\texttrademark} toolkit; for medium - ATSS head; and for accurate - VFNet with modified deformable convolutions (DCNv2 \cite{modifiedDCN}). 
In our experiments, as described later, we were not limited with these architectures to achieve specific goals of performance and accuracy. For example, we also tried lighted ATSS as a candidate for the fast model, and SSD with more heads as a candidate for the medium model. 
But the final setup with 3 different architectures worked best. 

\subsection{Training Pipeline}
\label{training_pipeline_section}

\paragraph{Pretraining weights}
To achieve fast model convergence and to get high accuracy right from the start, we used pretrained weights. 
Firstly, we trained MobileNetV2 classification model with ImageNet21k\cite{imagenet21k}, so a model could learn to differentiate trivial features and patterns. 
Afterward, we only took the trained MobileNetV2 backbone weight, cutting off the fully connected classifier head. 
Then, we used this pretrained backbone for our object detection model and trained ATSS-MobileNetV2 and SSD-MobileNetV2 on the COCO object detection dataset \cite{coco} with 80 annotated classes.
For VFNet we used pretrained COCO weights from OpenMMlab MMDetection \cite{openmmlab}.

\paragraph{Augmentation}
For augmentation, we used some classic tricks that have gained the trust of researchers. We augmented images with a random crop, horizontal flip, and brightness and color distortions.
The multiscale training was used for medium and accurate models to make them more robust.
Also, after a series of experiments we carefully and empirically chose certain resolutions for each model in order to meet the trade-off between accuracy and complexity. For example, SSD performed best on the square 864x864 resolution, ATSS on the rectangle (992x736), and for VFNet we used the default high resolution without any tuning (1344x800).
During resizing we did not keep the initial aspect ratio for 3 reasons:
\begin{enumerate}
\item to add extra regularization;
\item to avoid blank spaces after padding;
\item to avoid spending too much time on model resizing if the image aspect ratio differs a lot from the default.
\end{enumerate}
The impact of these tricks on accuracy and training time can be found in the Ablation study (section~\ref{ablation_section}).

\paragraph{Anchor clustering for SSD}
Despite the wide application of the SSD model, especially in production, its high inference speed, and good enough results, its dependence on careful anchor box tuning is a huge drawback.
It relies on the dataset and object size characteristics.
So, it becomes a challenge to build dataset-agnostic training using SSD model.
For that reason, we followed the logic in \cite{lighweight_model} and used information from the training dataset to modify the anchor boxes before training started.
We collect object size statistics and cluster them using KMeans algorithm \cite{kmeans} to find the optimal anchor boxes with the highest overlap among the objects. These modified anchor boxes are then passed for training. 
This procedure is used for each dataset and helps to make the light model more adaptive, improving the result without additional model complexity overhead.

\paragraph{Early Stopping and ReduceOnPlateau}
The complexity and size of the input dataset can vary drastically, making it hard to predict the number of epochs, needed to get a satisfying result and could often lead to large training times or even overfitting.
To prevent it, we used Early Stopping \cite{early_stop} to stop training after a few epochs if those did not improve the result further. Further, we used average precision (AP@[0.5,0.95]) on the validation subset as a target metric.
This problem with the unfixed number of epochs also creates uncertainty in determining learning rate schedule for the given scenario.
We utilized the adaptive ReduceOnPlateau \cite{reduceonplateau} scheduler in tandem with Early Stopping, as commonly employed, to circumvent this problem.
This scheduler reduces the learning rate automatically if the consecutive epochs do not lead to improvement in the target metric - average precision for the validation dataset in our case.
We implemented both of these algorithms for the MMDetection \cite{mmdetection} framework that forms the backbone of our codebase. It also powers the Intel Geti computer vision platform.
It is an OpenVINO{\texttrademark} toolkit fork of the OpenMMLab repository.

\paragraph{Iteration patience}
Another challenge is that the number of iterations in the epoch differs drastically from dataset to dataset depending on its length. This creates a problem to choose an appropriate "patience" parameter for Early Stopping and ReduceOnPlateau in case of the dataset-agnostic training.
For example, if we use the same patience for large datasets and small ones, the training with a small amount of data will stop too early and will suffer from undertraining.
A possible solution here would be to copy the training dataset several times for every training epoch, increasing the number of iterations in the epoch. 
However, this will also lead to large datasets suffering from overfitting and long training times.

To overcome this, we adapted the classic approaches of ReduceOnPlateau and Early Stopping by adding \textbf{iteration patience} parameter.
It works similarly to the epoch patience parameter but also ensures that a specific number of iterations were run during training on those epochs.

This approach increased the accuracy on the small datasets however it has some drawbacks as well.
The iteration patience could vary from architecture to architecture and must be chosen manually, which causes adding an extra hyperparameter. 
It leads to a less universal training pipeline and careful hyperparameter selection is required in case of trying new architectures, which could be time-consuming. 

Another caveat is that on small datasets a large part of the training cycle is still spent on the validation and checkpoint creation because of the short epochs. 
It leads to the fact that up to 50\% of the training is spent on not-so-useful checkpoints which are clearly undertrained and not really informative. 
Therefore, ideally, you should maximize the time spent on training compared to the time spent on validation and checkpoint creation.

\section{Experiments}
\subsection{Datasets}
For training purposes, we used 11 public datasets varying in terms of the domains, number of images, classes, object sizes, overall difficulty, and horizontal/vertical alignment.
BCCD\cite{bccd}, Pothole\cite{pothole}, WGISD1\cite{wgisd}, WGISD5\cite{wgisd} and Wildfire\cite{wildfire} datasets represent the category with the small number of train images. 
Here WGISD1 and WGISD5 share the same images, but WGISD1 annotates all objects as 1 class (grape), while WGISD5 classifies each grape variety.
Aerial\cite{aerial}, Dice\cite{dice}, MinneApple\cite{minneapple} and PCB\cite{pcb} represent datasets with small objects that average size is less than 5\% of the image.
The PKLOT\cite{pklot} and UNO\cite{uno} are large datasets with more than 4000 images. 

\begin{figure*}[t]
    \centering
    \includegraphics[width=18cm]{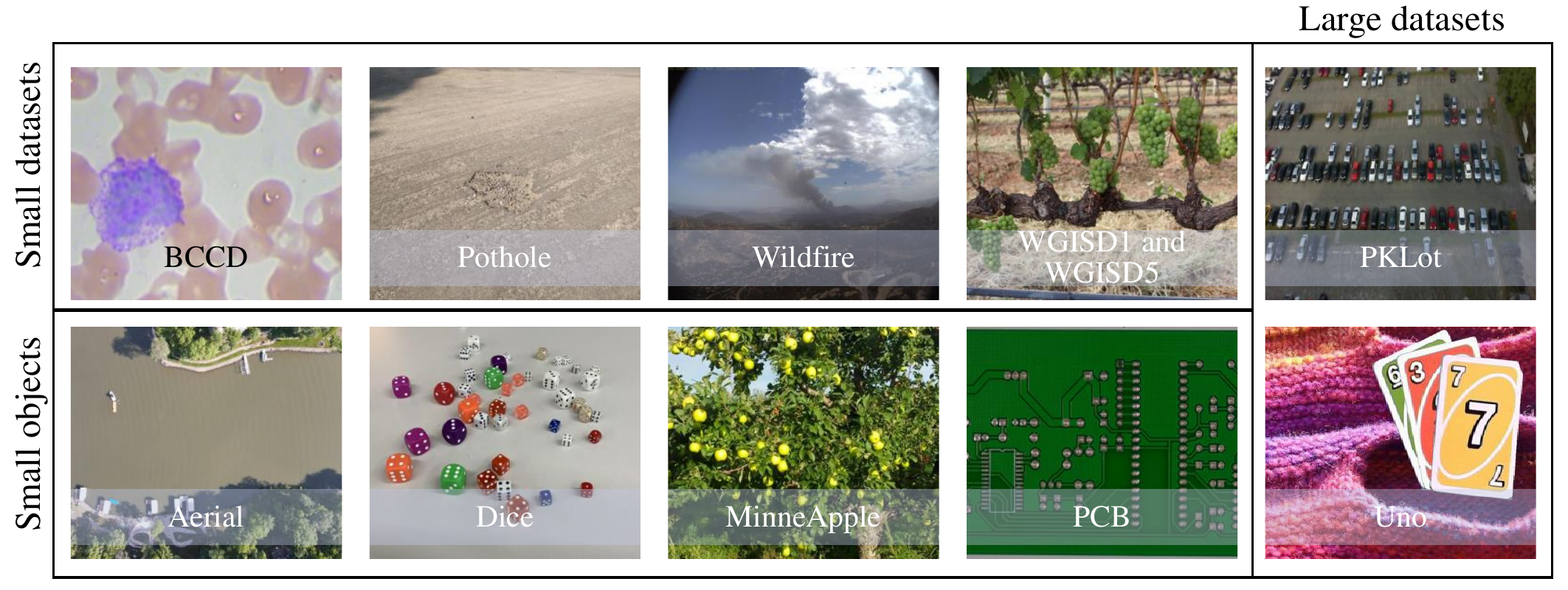}
    \caption{Samples of the Training Datasets.}
    \label{fig:data_sample}
\end{figure*}

The samples of the images can be found in Figure~\ref{fig:data_sample}. For illustration purposes, images were cropped to the square format.
The detailed statistics with splitting into sections can be found in Table~\ref{tab:data}.

\begin{table}[h!]
  \caption{Object Detection Training Datasets.}
  \label{tab:data}
  \centering
  \begin{tabular}{l|c|c|c|c|c}
    Dataset & Number of & Average object  &  \multicolumn{2}{c}{Number of Images} \\
            & classes & size, \% of image & Train     &   Validation \\ \hline

    BCCD & 3 & 17x21 & 255 & 73 \\

    Pothole & 1 & 23x14 & 465 & 133 \\
    WGISD1 & 1 & 7x16 & 110 & 27 \\
    WGISD5 & 5 & 7x16 & 110 & 27 \\
    Wildfire & 1 & 18x15 & 516 & 147 \\ \hline

    Aerial & 5 & 4x6 & 52 & 15 \\
    Dice & 6 & 5x3 & 251 & 71 \\
    MinneApple & 1 & 4x2 & 469 & 134 \\
    PCB & 6 & 3x3 & 333 & 222 \\ \hline

    PKLOT & 2 & 5x8 & 8691 & 2483 \\
    UNO & 14 & 7x7 & 6295 & 1798 \\
      \hline
    \hline
  \end{tabular}
\end{table}

\subsection{Evaluation Protocol}
For validation, we used a COCO Average Precision with IoU threshold from 0.5 to 0.95 (AP@[0.5,0.95]) as the common metric for the object detection task. 
\begin{displaymath}
\text{AP} = \sum_n (R_n - R_{n-1}) P_n
\end{displaymath}
where $P_n$ and $R_n$ are the precision and recall at the $n$-th threshold. AP averages 10 values because IoU thresholds belong to
[0.5,0.95] with an interval 0.05.

We collected train, val, and test AP for each dataset, evaluating the final weights on the train, validation, and test subset respectively. 
The val AP acted as the main metric for comparison.
To rank the different training strategies and architectures, the train, val, and test AP scores for each dataset were averaged across the data corpus.
Each of these AP scores contributes equally to the final metric without any weighting strategy.
As an auxiliary metric we used the same average AP scores but on different subsets:
\begin{itemize}
\item on datasets with small objects where average object size is less than 5\% of the image - to monitor how the network handles a hard task of the small object detection
\item on large datasets that contain more than 4000 images - to monitor how the network uses the potential of the large dataset and how its size impacts the training time.
\end{itemize}

\subsection{Results}

After more than 300 training rounds, we have chosen the 3 best architectures in terms of accuracy-performance trade-off: the lightweight and fast, the medium, and the heavy but highly accurate.
These are SSD on the MobileNetV2, ATSS on the MobileNetV2, VFNet on the ResNet50 respectively.
To increase the overall accuracy and keep training time within acceptable ranges we applied some common and architecture-aware practices, which are listed below and described in detail in Training pipeline section ~\ref{training_pipeline_section}.
The contribution of each of these techniques in both improving the accuracy and reducing the training time could be found in the Ablation study (section ~\ref{ablation_section}).

To choose the best models, we also experimented with well-known architectures in the scope of the lightweight, medium, and heavy models, such as YOLOX, FCOS, Faster R-CNN, and Cascade R-CNN.
We have tuned parameters, resolution, training strategy, and augmentation individually to achieve the best accuracy with comparable complexity.
The accuracy values of these architecture experiments could be found in Table~\ref{tab:avg_metrics}.

\begin{table*}[h!]
  \caption{Architecture Comparison.}
  \label{tab:avg_metrics}
    \centering
    \begin{tabular}{c|c|c|c|c|c}
      & & & \multicolumn{3}{c}{AVG AP on the} \\  
      \cline{4-6}
     Model &   AVG AP$^*$, \% & GFLOPs & small  & small object &  large  \\
     & & &  datasets,\% &  datasets, \% &   datasets,\%  \\

      \hline\hline
      \textit{Fast models}  \\
      YOLOX & 49.6 & 6.45 & \textbf{48.8} & 35.1 & 80.7 \\
     \textbf{SSD}  & \textbf{52.0} & 9.36 & 46.9 & \textbf{38.7} & \textbf{91.2} \\
      \hline\hline
     \textit{Medium models}     \\
      FCOS & 56.7 & 20.86 & 52.2 & 43.8 & 93.8  \\
     \textbf{ATSS} & \textbf{60.3} & 20.86 & \textbf{55.2} & \textbf{49.7} & \textbf{94.1}    \\
      \hline\hline
      \textit{Accurate models}     \\
      Faster R-CNN & 61.8 & 433.40 & 55.4& 53.7 & 94.0    \\
      Cascade R-CNN & 62.5 & 488.54 & 56.2&  55.0 & 93.6  \\
      \textbf{VFNet} & \textbf{64.7} &  347.78 & \textbf{59.5} & \textbf{56.1} & \textbf{95.2}   \\
      \hline \hline
      \multicolumn{5}{l}{\footnotesize{$^*$The average AP on the validation subsets was reported her}} \\
    \end{tabular}
  \end{table*}

For all of the models we used the following training setup: 
\begin{itemize}
\item start from the weights pretrained on the COCO;
\item augment images with a crop, flip, and photo distortions;
\item use ReduceOnPlateau learning rate scheduler with iteration patience;
\item use the Early Stopping to prevent overfitting on the large datasets and iteration  patience to prevent underfitting on the small ones.
\end{itemize}

Additional special techniques included:
\begin{itemize}
\item for SSD: anchor box reclustering and pretraining COCO weights on ImageNet-21k\cite{imagenet21k}
\item for ATSS: multiscale training and using ImageNet-21k weights
\item VFNet - multiscale training
\end{itemize}
Detailed results are in Table~\ref{tab:detailed_metrics} below.

\begin{table*}[h!]
  \caption{Detailed Results of the Chosen Models on All Datasets.}
  \label{tab:detailed_metrics}
    \centering

      \begin{tabular}{c|c|c|c|c|c|c|c|c|c|c|c} Model & \multicolumn{5}{c|}{Small datasets} & \multicolumn{4}{c|}{Datasets with small objects} & \multicolumn{2}{c}{Large datasets} \\ \cline{2-12}
      &

      \rotatebox[origin=c]{90}{BCCD} &	\rotatebox[origin=c]{90}{Pothole}	& \rotatebox[origin=c]{90}{WGISD1} &
      \rotatebox[origin=c]{90}{WGISD5} &	\rotatebox[origin=c]{90}{Wildfire} &

      \rotatebox[origin=c]{90}{Aerial} &	\rotatebox[origin=c]{90}{Dice} &	\rotatebox[origin=c]{90}{MinneApple}  &	\rotatebox[origin=c]{90}{PCB} &

      \rotatebox[origin=c]{90}{PKLOT}	&	\rotatebox[origin=c]{90}{UNO} \\
      \hline

      \hline\hline
      \textit{Fast models}  \\
      YOLOX & 63.5 & 44.7 & 44.1 & 47.7	& 47.7 & 24.2 &	58.2 & 23.0 & 34.9	&	80.9	&	80.5 \\
     \textbf{SSD}  & 60.0 & 37.5 & 45.9 & 40.2 & 51.0 & 22.7 & 61.3 & 31.1 & 39.8 & 94.7 & 87.7 \\
      \hline\hline
     \textit{Medium models}     \\
      FCOS & 63.5 & 42.4 & 55.2 & 55.0 & 44.7 & 29.3 & 59.8 & 39.5 & 46.7 & 97.3 & 90.2   \\
     \textbf{ATSS} & 63.5 & 47.9 & 57.5 & 55.4 & 51.5 & 36.8 & 73.8 & 41.0 & 47.3 & 97.7 & 90.5   \\
      \hline\hline
      \textit{Accurate models}     \\
      Faster R-CNN & 64.2 & 0.502 & 59.4	&	52.3 & 51.1	& 45.3 &	77.8 &	40.9 & 50.8	& 97.3 & 90.7  \\
      Cascade R-CNN & 65.5 & 51.2 & 60.5	&	53.2 & 50.4 & 45.9	& 79.5 & 42.7  & 51.7	& 95.9 & 91.3\\
      \textbf{VFNet} & 65.2 & 55.1 & 63.8 & 59.5 & 53.9 & 46.3 & 81.2 & 44.4 & 52.4 & 98.6 & 91.8 \\
      \hline \hline
      \multicolumn{12}{l}{\footnotesize{The average AP on the validation subsets was reported here}} \\
    \end{tabular}
  \end{table*}

\subsection{Ablation Study}
\label{ablation_section}
To estimate the impact of each training trick on the final accuracy, we conducted ablation experiments, by removing each of them from the training pipeline.
Based on our experiments, each of these tricks improved the accuracy of the target metric approximately by 1 AP
See results in Table~\ref{tab:ablation_accuracy}.
The iteration patience parameter showed great improvement (7.4 AP) for the fast SSD model, which stopped too early without the new parameter.
The best trick for medium ATSS is using COCO weights, which improved accuracy by 3.5 AP.
For the VFNet the most productive trick was to use COCO weights which increased the accuracy by 17 AP as well.

\begin{table*}[h!]
  \caption{Impact of Each Training Trick on the Accuracy of the Candidate Models.}
  \label{tab:ablation_accuracy}
  \centering
  \begin{tabular}{l|c|c|c|c}
    Configuration & SSD & ATSS & VFNet \\
    & AVG AP$^*$,\% & AVG AP,\% & AVG AP,\% \\
    \hline
    Final solution & \textbf{52.0} & \textbf{60.3} &  \textbf{64.7} \\
    w/o pretraining on the COCO &  51.2 & 56.8 & 47.7   \\
    w/o Iteration patience  & 44.6  & 58.6 & 64.3\\
    w/o ReduceOnPlateau LR scheduler & 50.9  & 58.9 &63.7  \\

    w/o Multicsale training & - & 59.2 & 64.0 \\

    \hline
    \textit{Architecture aware features} & & & \\
    w/o Anchor reclustering & 48.2 & - & - \\
    w/o modified DCN & - & - & 64.1 \\
    \hline
    \multicolumn{4}{l}{\footnotesize{$^*$The average AP on the validation subsets was reported here}} \\
  \end{tabular}
\end{table*}

Since we focused not only on the overall accuracy but also on getting results within an acceptable time range and fast inference of the trained model, we also used some tricks to reduce the training time and complexity of the model, while preserving the accuracy of predictions. 
We measured time spent on training in a single GPU setup, using GeForce 3090.
See Table~\ref{tab:ablation_training_time} for certain numbers.

The first thing it is seen here, which is counter-intuitive, is that the small SSD model averagely requires more epochs and more time to fulfill its potential and train enough than the medium ATSS model.
Talking about training time tricks, the most powerful was to use Early Stopping, which reduced the training time by 4.2-8.6 times.
Also, using the standard high resolution (1344x800) of the input images increased the training time for SSD and ATSS architectures by 25-40\%. However, for SSD architecture, the accuracy decreased by 0.9 AP due to the image resolution being too high for such a small model.
And, for ATSS, the accuracy increased by 1.7 AP, mostly due to datasets with small objects. But it slowed inference time by 1.5 times, rendering it too slow for our criteria for a medium model.

\begin{table*}[h!]
  \caption{Impact of Each Training Trick to the Training Time$^*$ of the Candidate Models.}
  \label{tab:ablation_training_time}
  \centering
  \begin{tabular}{l|c|c|c|c|c|c|c|c}
     & \multicolumn{2}{|c|}{SSD} & \multicolumn{2}{|c|}{ATSS} & \multicolumn{2}{|c|}{VFNet} \\

    Configuration & AVG training & AVG  & AVG training & AVG  & AVG training  & AVG  \\
    &  time, min &  epochs &  time, min &  epochs &  time, min & epochs  \\

    \hline
    Final solution & 45.18 & 145 & 36.27 & 92 & 230.91 & 71  \\
    w/o Early Stopping & 245.20 &300& 306.25 &300& 986.96&300 \\
    w/o Tuning resolution & 56.64 & 136 & 50.64 & 111 & - & - \\

    \hline
    \multicolumn{4}{l}{\footnotesize{$^*$GeForce RTX 3090 was used to benchmark training time.}}

  \end{tabular}

\end{table*}

\section{Conclusions}

In this research, we have outlined an alternative approach to training deep neural network models for fast-paced real-world use cases, across a variety of industries. This novel approach addresses the need for achieving high accuracy without performing time-consuming and computationally expensive training or fine-tuning for individual use cases. This methodology forms the backbone of the Intel Geti computer vision platform that enables rapidly building computer vision models for real world use cases.

As a first step, we have proposed a corpus of publicly available 11 datasets, spanning across various domains and dataset sizes.
Subsequently, we have performed more than 300 experiments, each consisting of parallel trainings -- one per each dataset in the selected corpus. We have selected average AP results over all datasets in the corpus as a target metric to optimize our final models.

Using this methodology, we have analyzed a number of architectures, concretely, SSD and YOLOX as fast model architectures for inference, ATSS and FCOS as medium model architectures, and VFNet, Cascade-RCNN, and Faster-RCNN as accurate models.
In the process, we have identified tricks and partial optimization that helped us optimize the average AP scores across the dataset corpus.

We encountered a significant challenge adjusting the optimal training time while building a universal template for a wide range of possible real-world use cases. This also affected the learning rate scheduler.
To solve this we introduced an additional iteration patience parameters for early stopping and ReduceOnPlateau along with the epoch patience parameter. This enabled us to balance the requirements for small and large datasets while maintaining optimal training times irrespective of the dataset sizes.

Our efforts result in 3 (one per 3 different performance-accuracy regimes) dataset-agnostic templates for object detection training, that provide a strong baseline on a wide variety of datasets and can be deployed on CPU using the OpenVINO{\texttrademark} toolkit.

\bibliographystyle{IEEEtran}
\bibliography{IEEEabrv, main}
%

\end{document}